\documentclass[10pt,twocolumn,letterpaper]{article}

\usepackage[pagenumbers]{iccv} 

\definecolor{iccvblue}{rgb}{0.21,0.49,0.74}
\usepackage[pagebackref,breaklinks,colorlinks,allcolors=iccvblue]{hyperref}

\title{Perception-as-Control: Fine-grained Controllable Image Animation with 3D-aware Motion Representation}

\author{Yingjie Chen, Yifang Men, Yuan Yao, Miaomiao Cui, Liefeng Bo\\
Institute for Intelligent Computing, Alibaba Tongyi Lab \\
{\tt\small \textcolor{red}{\href{https://chen-yingjie.github.io/projects/Perception-as-Control}{https://chen-yingjie.github.io/projects/Perception-as-Control}}}}

\begin{document}

\twocolumn[{
\maketitle
\begin{center}
    \captionsetup{type=figure}
    \vspace{-1.7em}
    \includegraphics[width=1.0\textwidth]{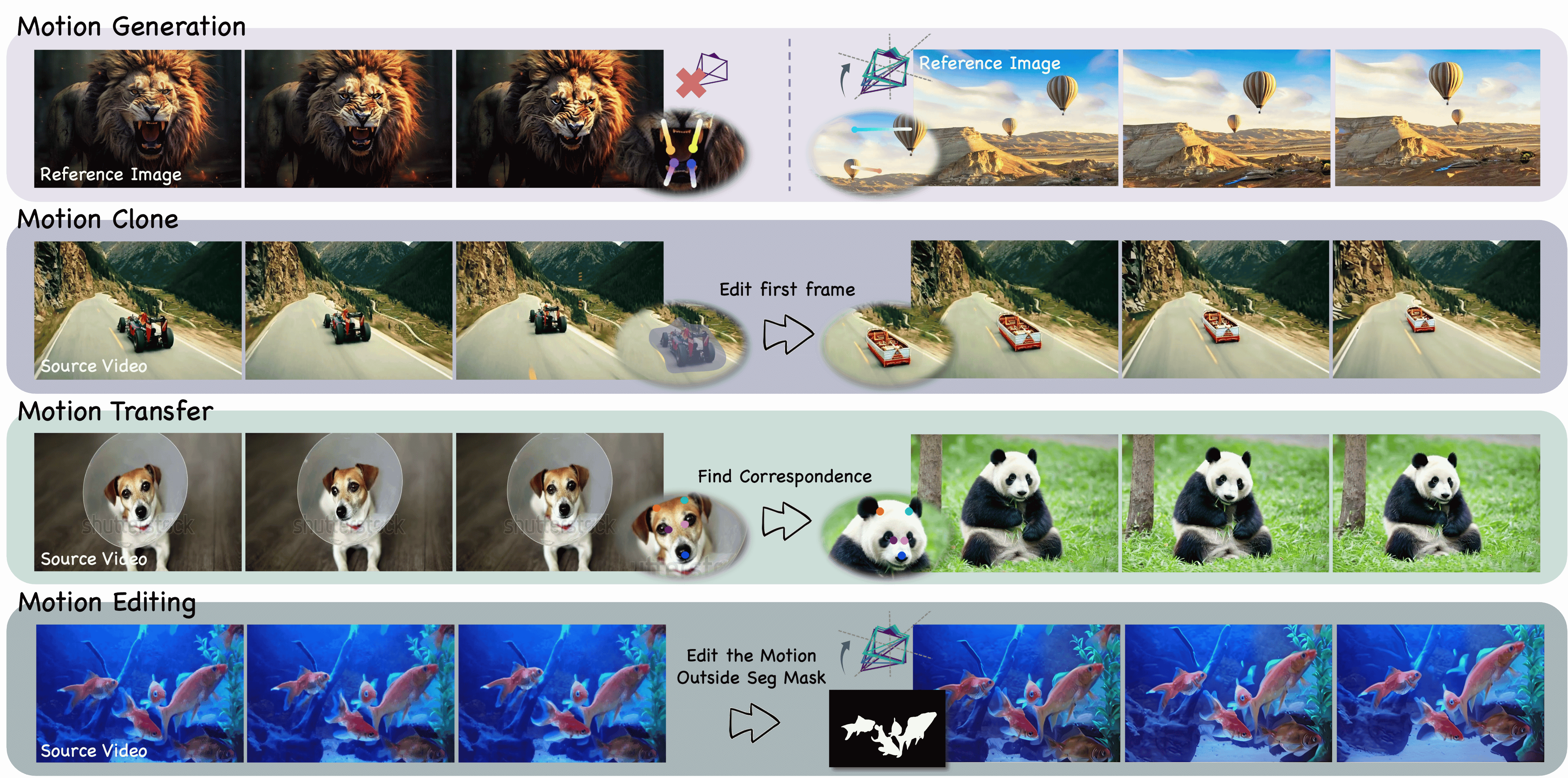}
    \captionof{figure}{Potential applications for Perception-as-Control. By constructing 3D-aware motion representation based on user intentions and utilizing the perception results as motion control signals, the proposed fine-grained motion-controllable image animation framework can be applied to various motion-related video synthesis tasks, such as \textbf{image-based} motion generation (animate image according to user instructions), and \textbf{video-based} motion clone (mimic the entire motions), motion transfer (relocate and rescale local motions based on semantic correspondence), local motion editing (edit fine-grained scene and object motions in user-specified regions).}
    \label{fig:teaser}
\end{center}
}]



\begin{abstract}
    Motion-controllable image animation is a fundamental task with a wide range of potential applications. Recent works have made progress in controlling camera or object motion via various motion representations, while they still struggle to support collaborative camera and object motion control with adaptive control granularity.
    To this end, we introduce 3D-aware motion representation and propose an image animation framework, called Perception-as-Control, to achieve fine-grained collaborative motion control.
    Specifically, we construct 3D-aware motion representation from a reference image, manipulate it based on interpreted user instructions, and perceive it from different viewpoints. In this way, camera and object motions are transformed into intuitive and consistent visual changes.
    Then, our framework leverages the perception results as motion control signals, enabling it to support various motion-related video synthesis tasks in a unified and flexible way.
    Experiments demonstrate the superiority of the proposed approach. For more details and qualitative results, please refer to our anonymous project webpage: \href{https://chen-yingjie.github.io/projects/Perception-as-Control}{Perception-as-Control}.
\end{abstract}

\vspace{-1em}
\section{Introduction}

With the proliferation of generative AI, image animation as a fundamental task is undergoing rapid progress~\cite{ma2024follow,guo2025sparsectrl,blattmann2023stable,yang2024cogvideox,niu2025mofa,li2024image,shi2024motion,wang2024levitor}. Various potential application scenarios put forward higher demands for this task, \ie, particularly in achieving precise motion-controllable image animation based on user intentions. 
Motion-controllable image animation usually takes a reference image and motion control signals derived from motion representation as input, aiming to generate videos with the expected motions while maintaining the specific appearance.
Since the reference image already includes rich appearance information, the key is to design better motion representations and effectively guide the motion content generation process.

Motion in videos can be decomposed into camera motion and object motion. Object motion refers to dynamic movements relative to the world coordinate system, while camera motion 
arises from viewpoint changes.
Although researchers have made great progress in designing various motion representations to obtain control signals for camera or object, achieving collaborative camera and object motion control with fine control granularity remains non-trivial. 
For camera motion control, some methods directly use camera intrinsic and extrinsic parameters~\cite{wang2024motionctrl} or convert them into Plücker embeddings~\cite{xu2024camco,he2024cameractrl,zheng2024cami2v} as high-level control signals, while others reconstruct point cloud to render incomplete frames~\cite{feng2024i2vcontrolcam,yu2024viewcrafter,hou2024training} or use dense optical flow~\cite{niu2025mofa,lei2024animateanything} as more concrete control signals. 
For object motion control, recent works usually follow an interactive setting that allows users to draw point trajectories on the reference frame. These methods convert sparse point trajectories into optical flow~\cite{wang2024motionctrl,li2024image,zhang2024tora,lei2024animateanything,niu2025mofa} or reorganize point embeddings~\cite{shi2024motion,wu2025draganything,geng2024motion,feng2024i2vcontrol} as control signals.
Nevertheless, works~\cite{lei2024animateanything,feng2024i2vcontrol,wang2024motionctrl,geng2024motion} focusing on collaborative motion control still suffer from conflict control issues and insufficient control granularity.

The key observations of this paper are three-fold: 
1) Motion representations based on pixel offsets such as optical flow are flexible for in-the-plane motions, while the lack of 3D awareness leads to conflict control issues with simultaneous camera and object motions.
2) Reconstruction-based motion representations such as point clouds show better spatial consistency, making them effective for camera motion control but inflexible to support fine-grained object motion control.
3) For controllable visual content generation tasks that maintain basic spatial structures, spatially aligned control signals are easier for alignment than high-level ones.
Based on these observations, an effective motion representation for fine-grained collaborative motion control should have the following properties.
First, the motion representation is equipped with 3D awareness to disentangle local components from the reference image.
Second, the motion representation is spatially consistent for stable camera motion control and flexible enough for fine-grained object motion control.
Third, the motion control signals derived from the motion representation are spatially aligned with the generated frames to enhance their connections.

\begin{figure}
    \centering
    \includegraphics[width=0.9\linewidth]{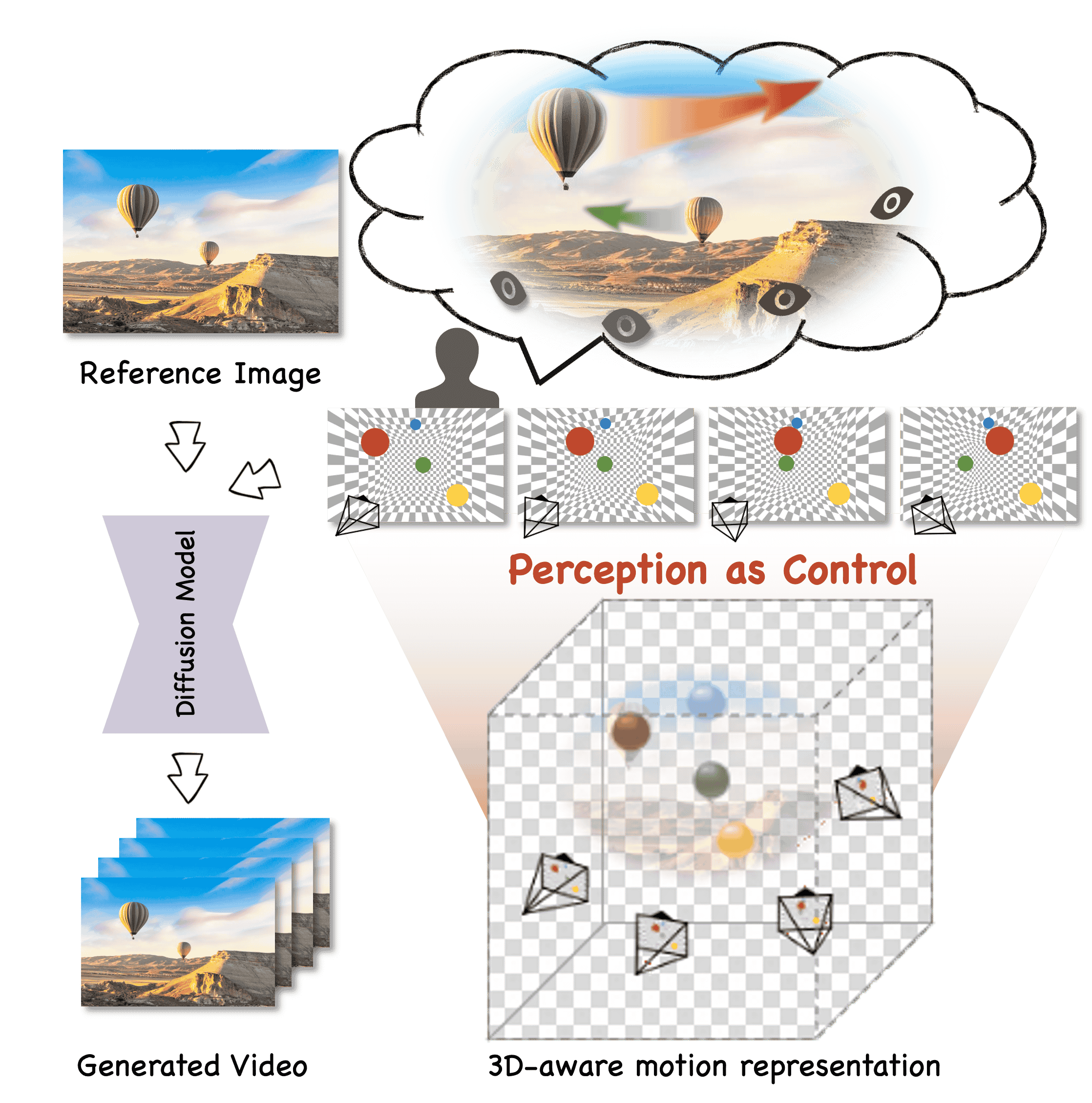}
    \caption{3D-aware motion representation for simplifying the 3D scene corresponding to the reference image, which consists of several unit spheres denoting key object parts and a world envelope accommodating them. We manipulate those spheres and observe the scene based on user intentions to transform camera and object motions into unified and consistent visual changes.}
    \label{fig:motivation}
\vspace{-1em}
\end{figure}

To this end, we introduce \textbf{3D-aware motion representation}.
In controllable image animation applications, users generally have clear intentions about how objects move and how to observe the scene.
The intuitive approach is to reconstruct the 3D scene based on the reference image, drive objects according to user intentions, and render the scene from the expected viewpoints.
Considering the significant effort required to reconstruct everything in the scene, we notice that human visual attention is predominantly drawn to moving parts, while the remaining parts are generally perceived as a whole~\cite{gibson2014ecological}.
Therefore, as shown in Fig.~\ref{fig:motivation}, we simplify and summarize the 3D scene as a combination of several unit spheres representing key object parts (most moving parts and a few static parts) and a vast world envelope akin to \emph{The Truman Show} accommodating them to demonstrate how the scene evolves as the observer’s viewpoint changes.
We then render the instantiated 3D scene to obtain perception results that reflect camera and object motions and necessary spatial relations.

Based on 3D-aware motion representation, we propose an image animation framework to achieve fine-grained collaborative motion control.
The framework takes a reference image and the perception results of the corresponding 3D-aware motion representation as motion control signals, which are spatially aligned with the generated frames.
Specifically, we apply two lightweight encoders to encode camera and object control signals separately to avoid RGB-level interference, and then merge them to form the final motion control signals.
We employ a U-Net-based diffusion model and inject both the reference image and motion control signals, enabling motion-controllable image animation.
A data curation pipeline is introduced to construct 3D-aware motion representations from in-the-wild videos. 
Moreover, we propose a three-stage training strategy to balance camera and object motion control during training.

Our contributions are three-fold:
\begin{itemize}
    \item We introduce 3D-aware motion representation to facilitate flexible and fine-grained collaborative motion control, which transforms camera and object motions into unified and consistent visual changes and is orthogonal to specific base model architectures.
    \item We propose an image animation framework to incorporate 3D-aware motion representation as spatially aligned motion control signals, which achieve collaborative motion control precisely and harmoniously.
    \item The proposed approach can support various motion-related applications via translating user intentions, such as motion generation, motion clone, motion transfer, and motion editing.
\end{itemize}

\section{Related Work}
Motion-controllable image animation has made significant progress recently. Since reference images have greatly limited the degree of freedom in generating contents, achieving fine-grained collaborative motion control is even harder for image animation than text-based video generation.

\noindent
\textbf{Camera Motion Control.} Works for camera motion control~\cite{wang2024motionctrl,xu2024camco,zheng2024cami2v,he2024cameractrl,feng2024i2vcontrolcam,yu2024viewcrafter,guo2025sparsectrl,sun2024dimensionx} aim to generate novel view images with spatial consistency. In computer graphics, camera information is typically represented using a set of intrinsic and extrinsic parameters. Methods like MotionCtrl~\cite{wang2024motionctrl} directly repeat these flattened camera parameters to image size as control signals.
CameraCtrl~\cite{he2024cameractrl} and CamCo~\cite{xu2024camco} transform the camera parameters into Plücker coordinates as pixel-wise embeddings, which is more discriminative. 
ViewCrafter~\cite{yu2024viewcrafter} and I2V-Camera~\cite{feng2024i2vcontrolcam} lift the reference image to 3D and render the point cloud to generate incomplete frames.
SparseCtrl~\cite{guo2025sparsectrl} and DimensionX~\cite{sun2024dimensionx} generate a specific set of training data for each type of camera motion and train individual LoRA models~\cite{hu2021lora} tailored for each camera motion pattern.
Concurrent work TrajectoryAttention~\cite{xiao2024trajectory} uses dense pixel trajectories to enhance temporal consistency, which is effective for camera motion control but hard to further support object motion control. 
Compared to high-level camera motion representations, our 3D-aware motion representation embodies camera motion information via visual changes of the world envelope, which is more intuitive and effective.

\noindent
\textbf{Object Motion Control.} Previous works usually follow the drag-based interaction setting for object motion control~\cite{yin2023dragnuwa,li2024image,shi2024motion,niu2025mofa,zhang2024tora,wu2025draganything}, which conveys user intentions via sparse point trajectories.
Methods such as DragNUWA~\cite{yin2023dragnuwa}, ImageConductor~\cite{li2024image}, Motion-I2V~\cite{shi2024motion} and MOFA-Video~\cite{niu2025mofa} employ optical flow as an intermediate motion representation. The former two methods transform point trajectories into sparse optical flow as control signals, while the latter two methods transform sparse point trajectories into dense optical flows via off-the-shelf or pretrained sparse-to-dense flow generation models.
Tora~\cite{zhang2024tora} enhances point trajectories into a trajectory map and employs 3D VAE to encode it.
DragAnyhting~\cite{wu2025draganything} obtains features for each pixel via DDIM inversion~\cite{song2020denoising} and reorganizes those features according to segmentation masks and point trajectories while being less effective in handling motions containing shape variations.
Although the above methods support camera control using point trajectories, they suffer from conflict control issues in local regions with simultaneous camera and object motions.
Instead of modifying 2D motion representation, we employ a simplified 3D scene and use its inherent perception results as control signals to avoid control conflicts. Due to this design, diffusion models can handle both local object motion and global camera motion simultaneously.

\noindent
\textbf{Collaborative Motion Control.} Only a few concurrent works~\cite{lei2024animateanything,geng2024motion,feng2024i2vcontrol} support collaborative motion control.
AnimateAnything~\cite{lei2024animateanything} applies two-stage generation, one for transforming camera and object control into optical flow representation, the other for generating videos based on that. However, they represent camera motion as Plücker embeddings, which is heterogeneous to optical flow and may lead to imprecise control.
MotionPrompting~\cite{geng2024motion} employs dense point embeddings constructed from 2D point tracking results as motion control signals, while such representation is not friendly enough for complicated motion control.
I2VControl~\cite{feng2024i2vcontrol} also employs dense point embeddings constructed via point transform and projection, while the granularity of object motion control is limited to object segmentation masks.
In contrast, we propose Perception-as-Control, which takes motion control signals derived from efficiently constructed 3D-aware motion representation and achieves fine-grained collaborative motion control.

\section{Perception-as-Control}

\begin{figure*}
    \centering
    \includegraphics[width=0.95\linewidth]{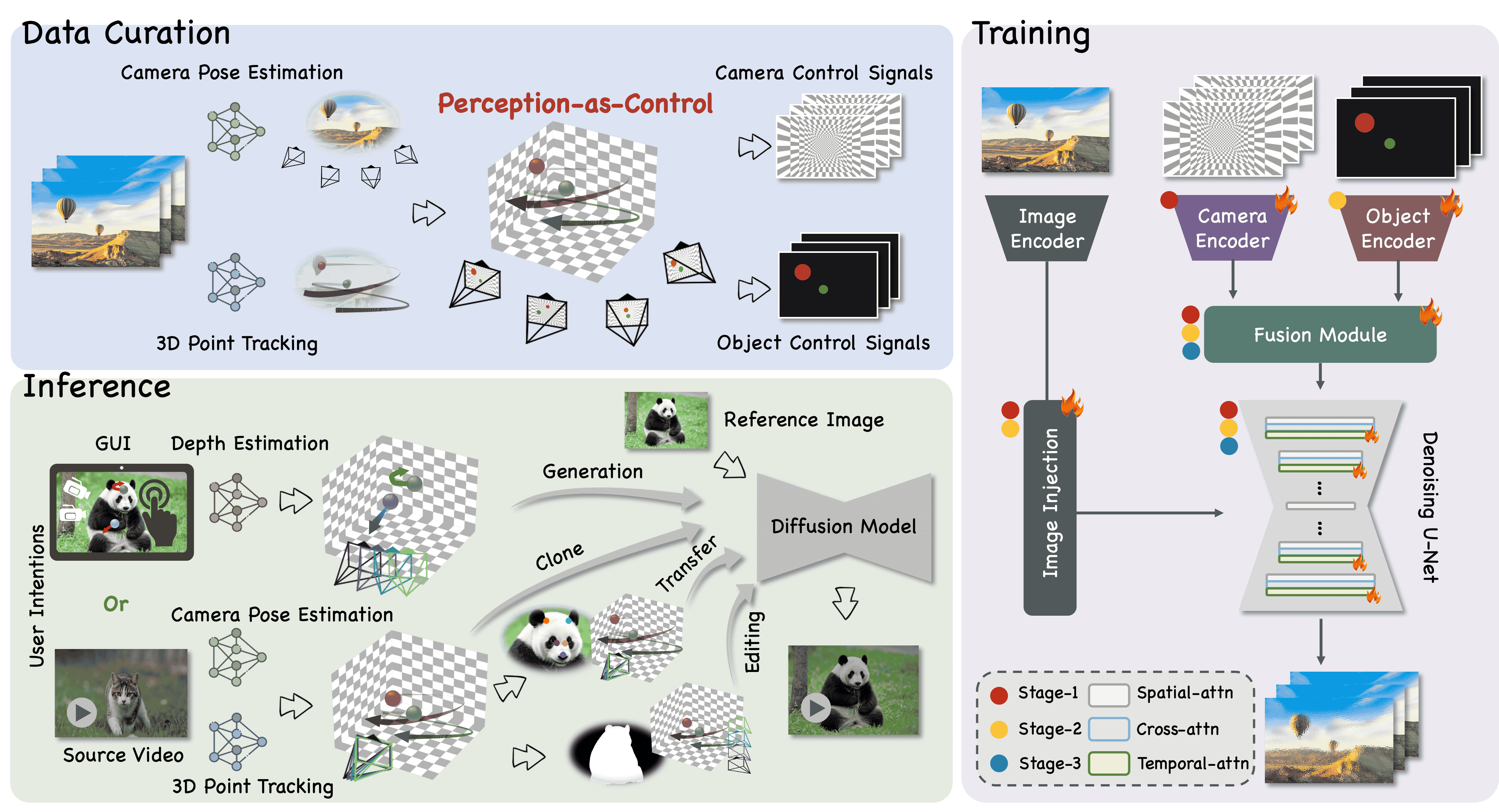}
    \vspace{-0.3em}
    \caption{Overview. In our data curation pipeline, an in-the-wild video is taken as input, and off-the-shelf algorithms are applied to obtain camera poses and 3D point locations. Based on the results, 3D-aware motion representation can be constructed and rendered to derive camera and object control signals. During training, those control signals are fed into separate encoders to avoid RGB-level interference and merged afterward. A three-stage training strategy is introduced to achieve fine-grained collaborative motion control. During inference, user intentions are first interpreted as 3D-aware motion representation, enabling our framework to support various motion-related applications.}
\label{fig:overview}
\vspace{-0.7em}
\end{figure*}

As shown in Fig.~\ref{fig:overview}, our approach consists of three primary parts: a data curation pipeline for constructing 3D-aware motion representation from in-the-wild videos, an image animation framework for motion-controllable video generation, and a three-stage training strategy to facilitate fine-grained collaborative motion control.

\subsection{Preliminaries}
Video Diffusion Model (VDM)~\cite{blattmann2023stable,guo2023animatediff,yang2024cogvideox} adopts the foundational principles of image diffusion probabilistic models and models the temporal dependencies between frames,
resulting in videos with consistent visual content and coherent motion.
Let $x_0\in\mathbb{R}^{L\times H\times W\times C}$ denote the initial video latent, where $L$ is the number of frames. The forward process is defined as a chain that progressively adds Gaussian noise to the original video, as shown in Eq.~\ref{eq:forward}:
\begin{equation}
\label{eq:forward}
    x_t = \sqrt{\hat{\alpha_{t}}x_{t - 1}} + \sqrt{{1 - \hat{\alpha_{t}}}\epsilon}, \epsilon \sim \mathcal{N}(\mathbf{0}, \mathbf{I}),
\end{equation}
where $t\in\{1, \dots, T\}$ denotes diffusion timestep, $\hat{\alpha_{t}}$ regulates the intensity of noise added at each $t$, and $\epsilon$ is drawn from standard Gaussian noise.
The optimization objective is to predict the probability distribution $p(x_{t - 1}|x_{t})$ using a parameterized network $\theta$. The loss function is in Eq.~\ref{eq:noise_prediction}:
\begin{equation}
\label{eq:noise_prediction}
    \mathcal{L}_{\theta} = \mathbb{E}_{x_0, \epsilon, \mathcal{C}, t}[||\epsilon - \hat{\epsilon_{\theta}}(x_t, \mathcal{C}, t)||^2_2],
\end{equation}
where $\mathcal{C}$ is an optional parameter denoting the conditions, such as reference images and motion control signals.

\subsection{3D-aware Motion Representation}
For training purposes, pairs of video clip and the corresponding 3D-aware motion representation are required. Unfortunately, commonly used large-scale video datasets predominantly consist of in-the-wild videos lacking camera settings and scene information.
Reconstructing a delicate 3D scene is computationally expensive and the scale ambiguity issue leads to non-negligible reconstruction errors.
Alternatively, constructing the simplified 3D scene directly via appropriate approximations can bypass these problems and is readily achievable even for in-the-wild videos.
Thus, we introduce our data curation pipeline, using off-the-shelf algorithms to construct 3D-aware motion representation for any collectible video.

\noindent
\textbf{SceneCut Detection and Motion Filtering.}
First, we apply common preprocessing steps to extract video clips containing continuous visual contents.
We employ PySceneDetect~\cite{pyscenedetect} to split videos into video clips.
Then, we use RAFT~\cite{teed2020raft} to estimate optical flow for each video clip and use the Frobenius norm as a motion score to filter out video clips with non-obvious motion (motion scores lower than $30^{\rm th}$ percentile), the same as previous methods~\cite{li2024image}.

\noindent
\textbf{3D Point Tracking for Local Object Motion.}
Key object parts refer to not only moving object parts but also stationary ones, which are critical for revealing spatial relations.
They are simplified as unit spheres in the 3D scene, and the number of these spheres is defined according to the control granularity flexibly.
We require 3D locations of key object parts at each moment to set these spheres initially and manipulate their movements. 
Therefore, we employ SpaTracker~\cite{xiao2024spatialtracker} to achieve 3D point tracking in each video clip. Note that combining algorithms of 2D point tracking~\cite{karaev2025cotracker} and depth estimation~\cite{piccinelli2024unidepth,yang2024depth,yang2024depthv2} also helps.
In our implementation, we track $25\times 25$ grid points for each video clip, resulting in 3D point tracking results $\mathbf{P}=[\mathbf{P}^{\rm x}, \mathbf{P}^{\rm y}, \mathbf{P}^{\rm z}]^{\mathsf{T}}\in \mathbb{R}^{L\times N\times 3}$ ($L$ is the number of frames and $N$ is the number of tracked points).

\noindent
\textbf{Camera Pose Estimation for Global Camera Motion.}
We employ a world envelope to mark the remaining parts, which is stationary relative to the world coordinate system.
To transform global camera motion into visual changes, we can use off-the-shelf algorithms~\cite{wang2021tartanvo,zhao2022particlesfm} to recover camera trajectories from a sequence of continuous frames and render the world envelope.
Despite the scale ambiguity problem, our world envelope, which serves as a scene marker, only requires coarse camera poses instead of extremely precise ones. 
In our implementation, we employ TartanVO~\cite{wang2021tartanvo} for this purpose, since it is fast and generalizes well across various cameras. Camera intrinsic parameters are set to $\mathbf{K}=[W, 0, W//2; 0, H, H//2; 0, 0, 1]$, where $W$ and $H$ are the width and height of the first frame.
For each video clip, we obtain the estimated camera poses $\mathbf{E}=\{\mathbf{E}_l|l\in \{1, 2 \dots, L\}\}$, where $L$ is the number of frames, $\mathbf{E}_l=[\mathbf{R}\in\mathbb{R}^{3\times 3}, \mathbf{t}\in\mathbb{R}^{1\times 3}]$ are camera extrinsic parameters at frame $l$.

\noindent
\textbf{Perceptual 3D Scene Rendering.}
Once we have obtained the 3D point tracking results $P$ and the estimated camera poses $E$, we construct our 3D-aware motion representation and render it as motion control signals.

First, we set $N$ unit spheres with different colors representing key object parts.
These spheres are rendered on the pixel plane according to a specific camera pose and appear as circles that are larger near and smaller far away in the rendering images under perspective projection.
Since the estimated depth is not precise enough in some cases, we normalize point depths $\mathbf{P}^{\rm z}$ as $\mathbf{P}^{\rm d}$ for each video clip to only focus on their relative distance.
After that, we set the projected circle radius of the $n^{\rm th}$ sphere to $1 - \mathbf{P}^{\rm d}_{l, n}$ ($\mathbf{P}^{\rm d}_{l, n}$ is the normalized depth of the $n^{\rm th}$ sphere at frame $l$).
For convenience, we generate a continuous color map where each pixel has a distinct color. We then set the color of each sphere to the corresponding color of its projection center in the first frame.
Finally, these spheres are projected and rendered as colorful circles with varying radii.

Then, we construct a cube centered at the origin with the side length of $z_{\rm far}$ ($z_{\rm far}$ is the far clipping plane of a viewing frustum) as an implementation of our world envelope. 
We apply checkerboard textures to increase the recognizability of the world envelope, which helps demonstrate how the scene evolves as the viewpoint changes.
We align the camera coordinate system to the world coordinate system for the first frame and render the subsequent frames according to the relative camera pose sequence $E$, without loss of generality.
\emph{Note that the implementation of the world envelope is not limited to a cube, and its textures can vary beyond the checkerboard style.}
To avoid RGB-level interference, we render unit spheres and world envelope separately as two spatial layers, acting as object and camera control signals respectively.

\subsection{Network Architecture}
We propose an image animation framework to incorporate our 3D-aware motion representation, which takes a reference image and a sequence of spatially aligned motion control signals as input and generates a video as output.

As shown in Fig.~\ref{fig:overview}, our framework is based on a denoising U-Net architecture, with extra motion modules~\cite{guo2023animatediff} for modeling temporal information.
Under the image animation setting, it is essential to preserve the appearance information contained in the reference image (image control signals) and follow the motion information contained in motion control signals. 
Previous methods inject control signals by directly adding them to the input noise~\cite{ma2024followpose,men2024mimo} or integrating them into denoising U-Net through auxiliary architectures~\cite{zhang2023adding,ye2023ip,ma2024followclick,hu2024animate,wang2024motionctrl,he2024cameractrl,niu2025mofa}.
Given that the motion control signals are already spatially aligned, in our framework, we inject motion control signals by adding them to the noise and inject the reference image via ReferenceNet~\cite{hu2024animate} for better appearance preservation. 
Specifically, we use two lightweight encoders to encode camera and object control signals separately, avoiding RGB-level interference, and then use a fusion module to merge them as the final motion control signals. We implement the fusion module empirically as a convolution block.

During training, training data are prepared through our data curation pipeline, and a three-stage training strategy is applied for balancing camera and object motion control. During inference, user intentions conveyed in various forms are transformed into motion control signals via 3D-aware motion representation, as the input of our framework.

\subsection{Training Strategy}
Although both camera and object motions can be transformed into unified and consistent visual changes through 3D-aware motion representation, video clips with entangled motions increase the difficulty of training.

\noindent
\textbf{Stage 1: Camera Motion Control Training.}
In the first stage, we use video clips from RealEstate-10K\cite{zhou2018stereo} with camera motion only to train our camera encoder.
We temporarily exclude object encoder and fusion module from the framework.
Since pure viewpoint changes bring about coherent and consistent overall scene movements, we train motion modules together to capture the correspondence between changes of the world envelope and scene contents.
Thus, we optimize camera encoder, motion modules, and ReferenceNet in this stage.

\noindent
\textbf{Stage 2: Collaborative Motion Control Training.}
In the second stage, we add video clips from WebVid-10M\cite{bain2021frozen} that contain both camera and object motions for training.
We add object encoder and fusion module back for collaborative motion control.
Motion modules are optimized to align each rendered unit sphere with its corresponding object parts, since the initial location of each unit sphere is not restricted.
To speed up the alignment between unit spheres and object parts, we use dense unit spheres in this stage.
We fix camera encoder and optimize object encoder, fusion module, motion modules, and ReferenceNet.
Moreover, to preserve the effectiveness of both control signals, especially in conflict cases, we randomly drop one type of control signal at a fixed rate for RealEstate10K video clips. 

\noindent
\textbf{Stage 3: Dense-to-Sparse Fine-tuning.}
This stage aims to achieve fine-grained motion control with our world envelope and sparse unit spheres, which requires the model to determine each sphere's control range adaptively.
Considering that object motion tends to occur on salient objects, we employ an off-the-shelf segmentation algorithm~\cite{liu2020boosting} to obtain salient objects in the reference image and select unit spheres inside their mask as \emph{Set1}.
Then we calculate the trajectory length of each unit sphere and select those with a trajectory length higher than the $80^{\rm th}$ percentile as \emph{Set2}.
By combining \emph{Set1} and \emph{Set2}, we obtain the dense sphere set and randomly sample $N\in \{1, 2, \dots, 16\}$ spheres from it per iteration. In this stage, we fix all components except fusion module and motion modules during training.

For all stages, the loss function is in Eq.~\ref{eq:loss_function}:
\begin{equation}
    \mathcal{L} = \mathbb{E}_{x_0, c_{\rm img}, c_{\rm cam}, c_{\rm obj}, t}[||\epsilon - \epsilon_{\theta}(x_t, c_{\rm img}, c_{\rm cam}, c_{\rm obj}, t)||^2_2],
\label{eq:loss_function}
\end{equation}
where $\epsilon \sim \mathcal{N}(\mathbf{0}, \mathbf{I})$ is drawn from standard Gaussian noise, $t\in\{1, \dots, T\}$ denotes the diffusion timestep, and $c_{\rm img}$, $c_{\rm cam}$, $c_{\rm obj}$ are the reference image, camera motion control signals and object motion control signals, respectively.

\begin{figure}[!t]
    \centering
    \includegraphics[width=1.0\linewidth]{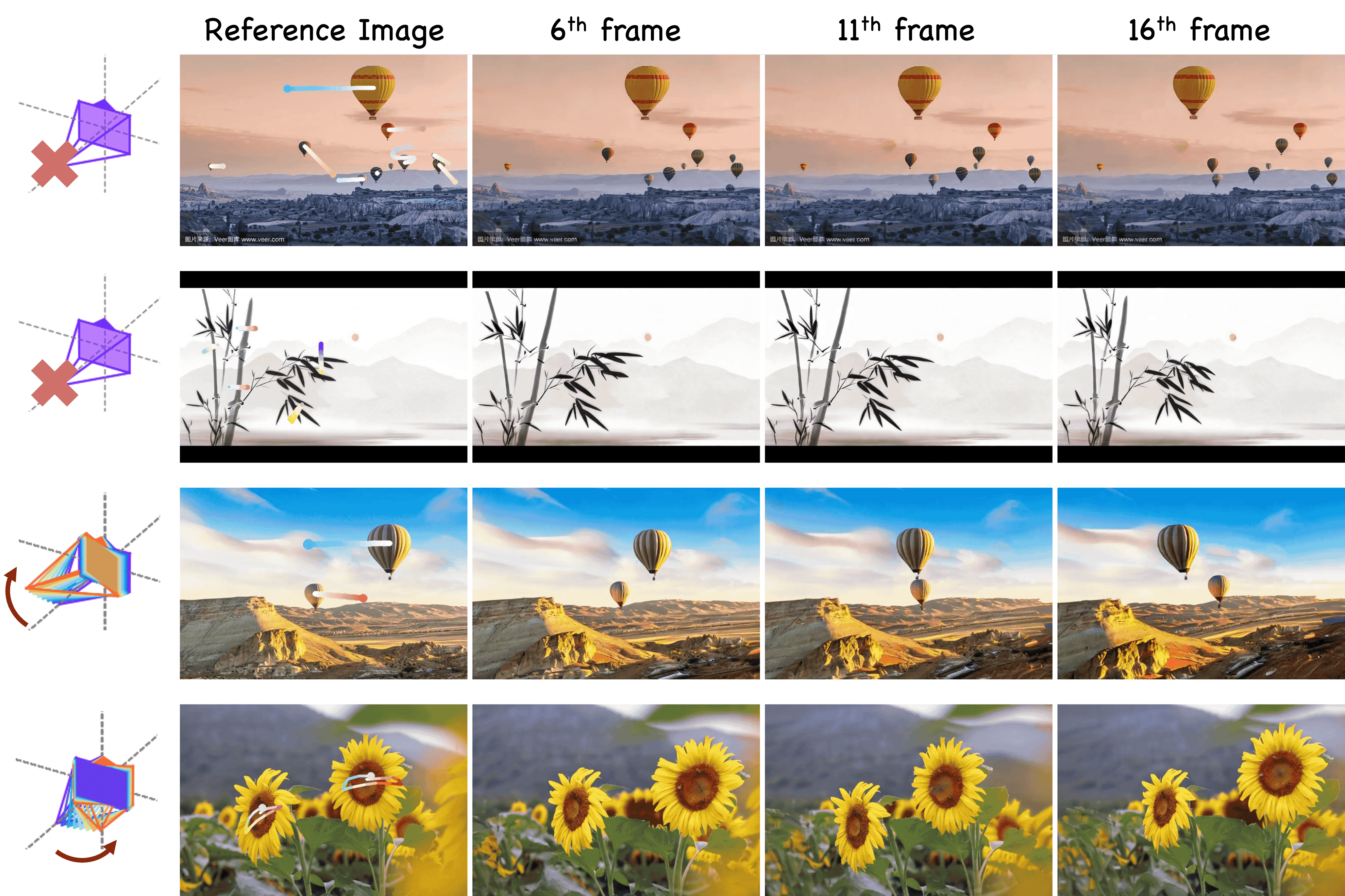}
    \caption{Results of fine-grained collaborative motion control.}
    \label{fig:results}
\end{figure}

\section{Experiments}

\begin{table}[!t]
\scriptsize
\centering
\begin{tabular}{c|cccc}
    \toprule
    & FID$\downarrow$ & FVD$\downarrow$ & RotErr$\downarrow$ & TransErr$\downarrow$ \\
    \midrule
    CameraCtrl & 14.538 & 178.497 & 0.648 & 0.203 \\
    Ours$_{14}$ & \textbf{11.795} & \textbf{73.449} & \textbf{0.641} & \textbf{0.199} \\
    \midrule
    MotionCtrl & 17.081 & 122.570 & 0.957 & 0.278 \\
    CameraCtrl* & 14.942 & 136.602 & 0.837 & 0.257 \\
    \midrule
    MotionCtrl++ & 16.327 & 104.020 & 0.906 & 0.265 \\ 
    CameraCtrl*++ & 12.923 & 113.165 & 0.761 & 0.291 \\
    Ours & \textbf{9.675} & \textbf{52.421} & \textbf{0.757} & \textbf{0.235} \\
    \bottomrule
\end{tabular}
\caption{Comparison on RealEstate test set for camera motion control. CameraCtrl* denotes the 16-frame version we reproduced.}
\label{tab:cmp_sota_cam}
\end{table}

\begin{table}[!t]
\scriptsize
\centering
\begin{tabular}{c|cccc}
    \toprule
    & FID$\downarrow$ & FVD$\downarrow$ & ObjMC$\downarrow$ & SubCon$\uparrow$ \\
    \midrule
    Motion-I2V & 60.643 & 367.524 & 34.52 & 0.845 \\
    MOFA-Video & 26.895 & 175.701 & 30.26 & 0.944 \\
    \midrule
    MotionCtrl++ & 32.528 & 208.387 & 28.57 & 0.959 \\
    CameraCtrl*++ & 28.393 & 164.654 & 25.72 & 0.963 \\
    Ours & \textbf{16.465} & \textbf{161.076} & \textbf{23.32} & \textbf{0.968} \\
    \bottomrule
\end{tabular}
\caption{Comparison on WebVid test set for object motion control.}
\label{tab:cmp_sota_obj}
\vspace{-2em}
\end{table}

\begin{figure*}[!t]
    \centering
    \includegraphics[width=1.0\linewidth]{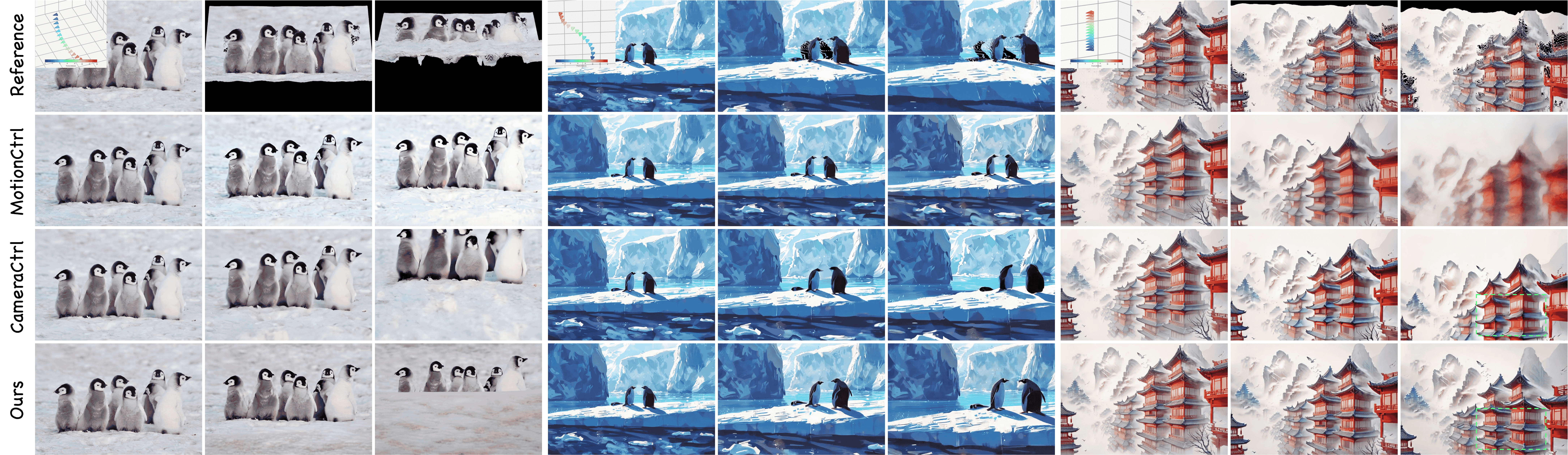}
    \caption{Qualitative comparison with state-of-the-art methods regarding camera motion control. \emph{Reference} images are generated by reconstructing point clouds based on the reference image and rendering them according to camera parameters.}
    \label{fig:cmp_sota_cam}
\end{figure*}

\begin{figure*}[!t]
    \centering
    \includegraphics[width=1.0\linewidth]{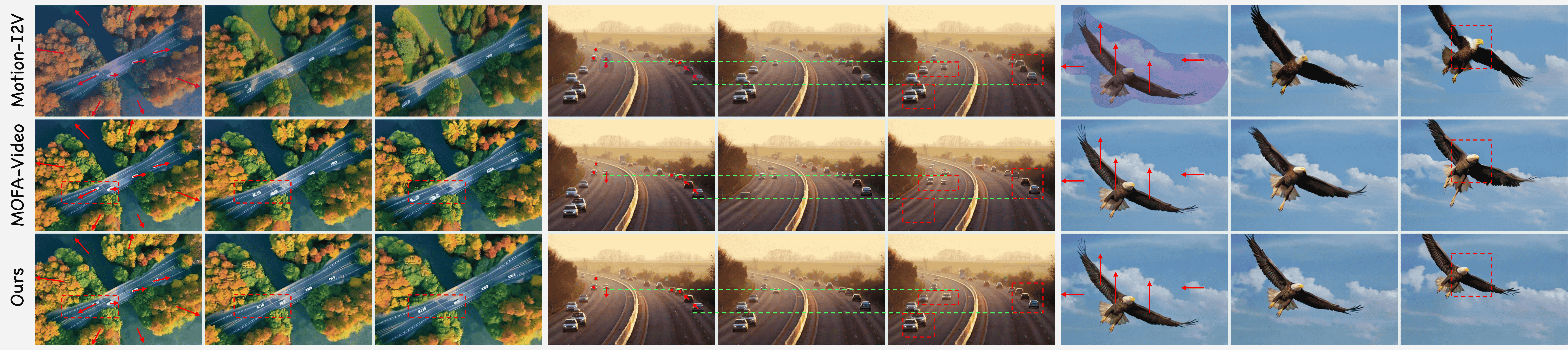}
    \caption{Qualitative comparison with state-of-the-art methods regarding object motion control and simple collaborative motion control.}
    \label{fig:cmp_sota_obj}
\end{figure*}

\subsection{Experimental Settings}
\noindent
\textbf{Datasets.}
We use video datasets RealEstate-10K~\cite{zhou2018stereo} and WebVid-10M~\cite{bain2021frozen} for training. RealEstate-10K contains videos with pure camera motion, and each video is annotated with ground-truth camera parameters. WebVid-10M is a large-scale dataset containing in-the-wild videos with both camera and object motions. After our data curation pipeline, the training data contains around 61K video clips from RealEstate and 352K from WebVid.

\noindent
\textbf{Metrics.}
We adopt Frechet Video
Distance (FVD)~\cite{unterthiner2018towards}, Frechet Inception Distance (FID)~\cite{heusel2017gans} to measure visual quality.
ObjMC~\cite{wang2024motionctrl} and Subject Consistency (SubCon) in VBench~\cite{huang2024vbench} are used to measure object motion controllability and average rotation error (RotErr) and translation error (TransErr)~\cite{he2024cameractrl} are used to measure camera motion controllability.
We select 1K clips from RealEstate test set and 1K from WebVid test set for quantitative evaluation.

\noindent
\textbf{Implementation Details.}
We adopt Stable Diffusion 1.5 (SD1.5)~\cite{rombach2022high} as our base model, and initialize both denoising U-Net and ReferenceNet~\cite{hu2024animate} using its pretrained weights. We add motion modules as in previous works~\cite{guo2023animatediff,blattmann2023stable} with random initialization. During training, CLIP~\cite{ramesh2022hierarchical} image encoder and VAE~\cite{kingma2013auto} encoder and decoder remain frozen. The model generates videos with 16 frames at a resolution of $768\times 512$. We use AdamW~\cite{loshchilov2017decoupled} as optimizer. We conduct the training on one NVIDIA A100 GPU. We train around 20k iterations for both stage 1 and stage 2 with learning rate of 1e-5, and 50k for stage 3 with learning rate of 1e-6, with a batch size of 1 for convergence.

\begin{figure}[!b]
    \centering
    \includegraphics[width=0.9\linewidth]{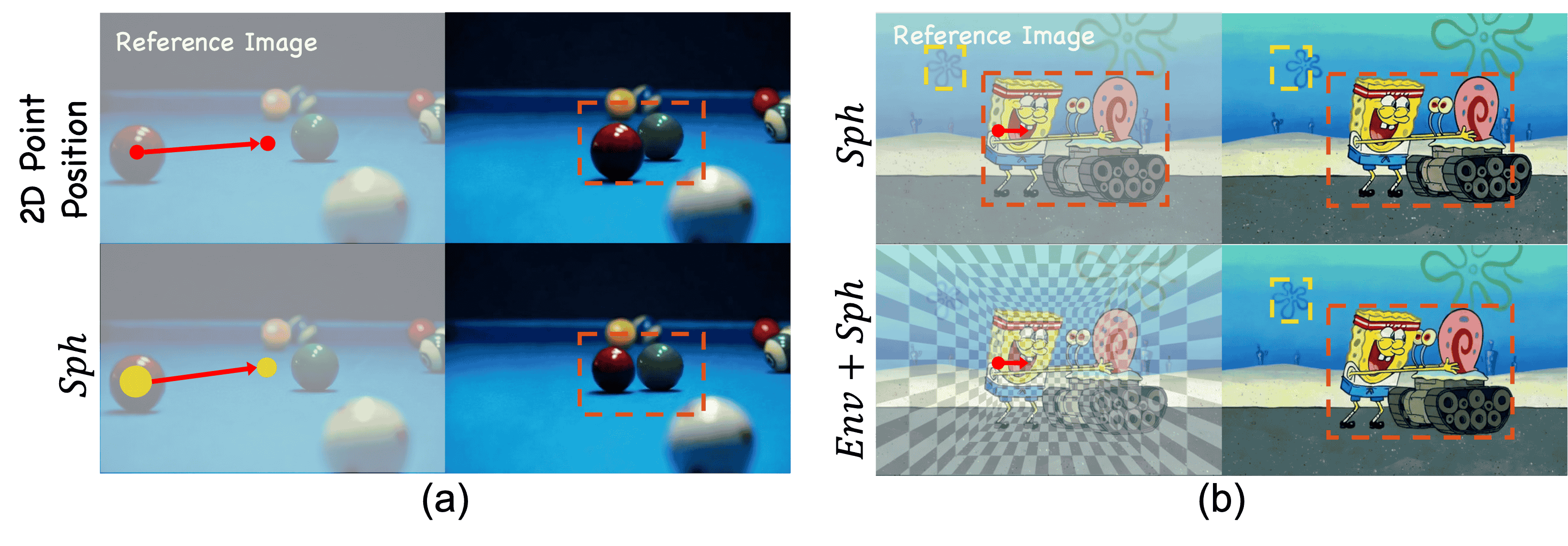}
    \vspace{-1em}
    \caption{Qualitative evidence for ablation study.}
    \label{fig:ablation}
\end{figure}

\begin{table}[!b]
\scriptsize
\centering
\begin{tabular}{c|cccc}
    \toprule
    & FID$\downarrow$ & FVD$\downarrow$ & ObjMC$\downarrow$ & SubCon$\uparrow$ \\
    \midrule
    2D point offset & 26.823 & 254.907 & 53.28 & 0.967 \\
    2D point position & 16.520 & 177.710 & 30.61 & 0.968 \\
    $Sph$ (3D) & \textbf{16.458} & 176.989 & 29.47 & 0.968 \\
    $Env$ + $Sph$ (Ours) & 16.465 & \textbf{161.076} & \textbf{23.32} & \textbf{0.969} \\
    \bottomrule
\end{tabular}
\caption{Ablation study on 3D-aware motion representation. $Env$ denotes our world envelope and $Sph$ denotes 3D unit spheres.}
\label{tab:ablation_obj}
\end{table}

\begin{table}[!t]
\scriptsize
\centering
\begin{tabular}{c|cccc}
    \toprule
    & FID$\downarrow$ & FVD$\downarrow$ & ObjMC$\downarrow$ & SubCon$\uparrow$ \\
    \midrule
    One-stage & 59.040 & 318.941 & 36.72 & 0.959 \\
    Two-stage & 26.823 & 254.907 & 31.29 & 0.966 \\
    Three-stage (Ours) & \textbf{16.465} & \textbf{161.076} & \textbf{23.32} & \textbf{0.969} \\
    \bottomrule
\end{tabular}
\caption{Ablation study on the proposed training strategy.}
\vspace{-2em}
\label{tab:ablation_strategy}
\end{table}

\begin{figure*}[!t]
    \centering
    \includegraphics[width=0.98\linewidth]{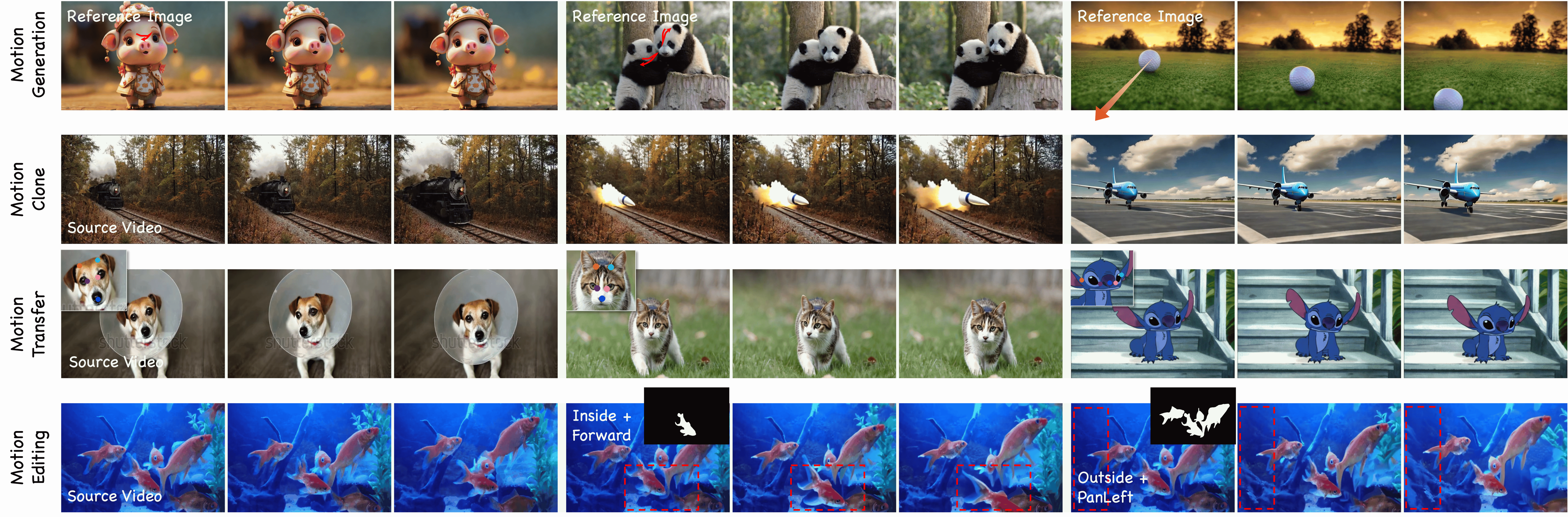}
    \caption{Results on four kinds of motion-related applications. More details and results please refer to Supplementary Materials.}
    \label{fig:applications}
\end{figure*}

\subsection{Comparison with State-of-the-Art Methods}
We compare our approach with state-of-the-art methods in Tab.~\ref{tab:cmp_sota_cam} and Tab.~\ref{tab:cmp_sota_obj}. We use the SVD version of MotionCtrl~\cite{wang2024motionctrl} and CameraCtrl~\cite{he2024cameractrl} for image-based generation. MotionCtrl directly takes camera extrinsic parameters as control signals, and CameraCtrl uses Plücker embeddings. Motion-I2V~\cite{shi2024motion} and MOFA-Video~\cite{niu2025mofa} employ dense optical flow as the final control signals. 
Since the proposed 3D-aware motion representation is orthogonal to specific base models, we also incorporate our representation into CameraCtrl and MotionCtrl and train them following our training strategy to empower them with fine-grained collaborative motion control capability, denoted as CameraCtrl++ and MotionCtrl++, respectively (see Sec. B in Supp.). By incorporating our 3D-aware motion representation in CameraCtrl and MotionCtrl, their ++ versions perform better on camera motion control and are capable of collaborative motion control. Moreover, ours outperforms their ++ versions, which shows the superiority of our framework.

Our approach achieves the best performance in terms of both camera and object motion control, as further evidenced in Fig.~\ref{fig:cmp_sota_cam} and Fig.~\ref{fig:cmp_sota_obj}.
In Fig.~\ref{fig:cmp_sota_cam}, we notice that CameraCtrl degrades on out-of-distribution camera trajectories such as large orbitary ones and tends to lose details in later frames, and MotionCtrl fails to adhere to the designated camera trajectories strictly and causes image blurring under large camera motions. In contrast, our approach generalizes well and achieves precise camera motion control.
From Fig.~\ref{fig:cmp_sota_obj} we can observe that optical flow-based methods face conflict control issues in collaborative motion control and struggle to maintain object appearance under large motions.

\subsection{Ablation Study}
We conduct ablation studies to demonstrate the effectiveness of 3D-aware motion representation and validate the necessity of our three-stage training strategy.
In Tab.~\ref{tab:ablation_obj}, \emph{2D point offset} refers to motion representations in previous works~\cite{wang2024motionctrl,li2024image}, and \emph{2D point position} is similar to visualizing the projection of unit sphere centroids with a fixed radius. We have these observations: 1) \emph{2D point offset} underperforms others, and we argue that this form is neither as complete as dense optical flow nor visually aligned as point positions, making it hard to associate points with object parts in the reference image. 2) The lack of depth information in \emph{2D point position} makes the spatial relations less clear and harms performance (Fig.~\ref{fig:ablation}a). 3) Only using \emph{3D unit spheres} may cause the ambiguity problem (Fig.~\ref{fig:ablation}b). 

In Tab.~\ref{tab:ablation_strategy}. \emph{One-stage} refers to directly optimizing all the modules mentioned using mixed data and \emph{Two-stage} refers to merging the last two stages into one. The results show that our carefully designed training strategy is essential in reducing training difficulty and achieving harmonious collaborative motion control (See Sec. C in Supp.). 

\subsection{Fine-grained Collaborative Motion Control}
Fig.~\ref{fig:results} shows the superiority of Perception-as-Control on fine-grained collaborative motion control.
For clarity, we project unit sphere centroids onto the reference image to visualize how they change over time (colors represent movement direction).
In the first row, we set object motions for several hot air balloons, and the result precisely reflects each one. In the second row, when setting several unit spheres to the bamboo simultaneously, our approach adaptively determines the control range for each one.
In the last two rows, we set both camera and object motions, the results adhere to the camera motion and present the adapted object motions (see Sec. A in Supp.).

\subsection{Applications}

As illustrated in Fig.~\ref{fig:teaser}, our approach has the potential to support various motion-related applications.
More results are given in Fig.~\ref{fig:applications}. 
For \textbf{motion generation}, we construct 3D-aware motion representation based on the reference image and manipulate it according to user-provided 2D/3D trajectories. 
Our approach handles motions including natural swaying, shape changes, intersections, and large displacements.
For \textbf{motion clone}, we construct 3D-aware motion representation from the source video as motion control signals, and edit the first frame as the reference image. The results exactly mimic the entire motions (camera and object) in the source video.
For \textbf{motion transfer}, we extract 3D trajectories of unit spheres corresponding to semantic points and relocate them to match those in the reference image. The results show that local motions from the source video are successfully transferred to objects that differ in scale and location.
For \textbf{motion editing}, we partially manipulate the 3D-aware motion representation based on user-specified regions, \eg, reset unit sphere trajectories or viewpoint changes. The results demonstrate the flexibility of our framework (see Sec. D in Supp.).



\section{Conclusion}
This paper aims to achieve precise and flexible fine-grained collaborative motion control, unlike previous methods that only support coarse-grained or inflexible collaborative motion control.
We observe the defects of previous motion representations and tackle the problems by proposing 3D-aware motion representation. This representation simplifies the corresponding 3D scene by transforming camera and object motions into unified and consistent visual changes.
The proposed image animation framework takes motion control signals derived from 3D-aware motion representation and can support various motion-related applications. 
Our extensive experiments show that our 3D-aware motion representation is orthogonal to model architectures and improves motion controllability consistently on multiple benchmarks (see Sec. E in Supp. for limitations).

{\small
\bibliographystyle{ieeenat_fullname}
\bibliography{main}
}


\newpage


\maketitlesupplementary

\begin{figure*}[!t]
    \centering
    \includegraphics[width=1.0\linewidth]{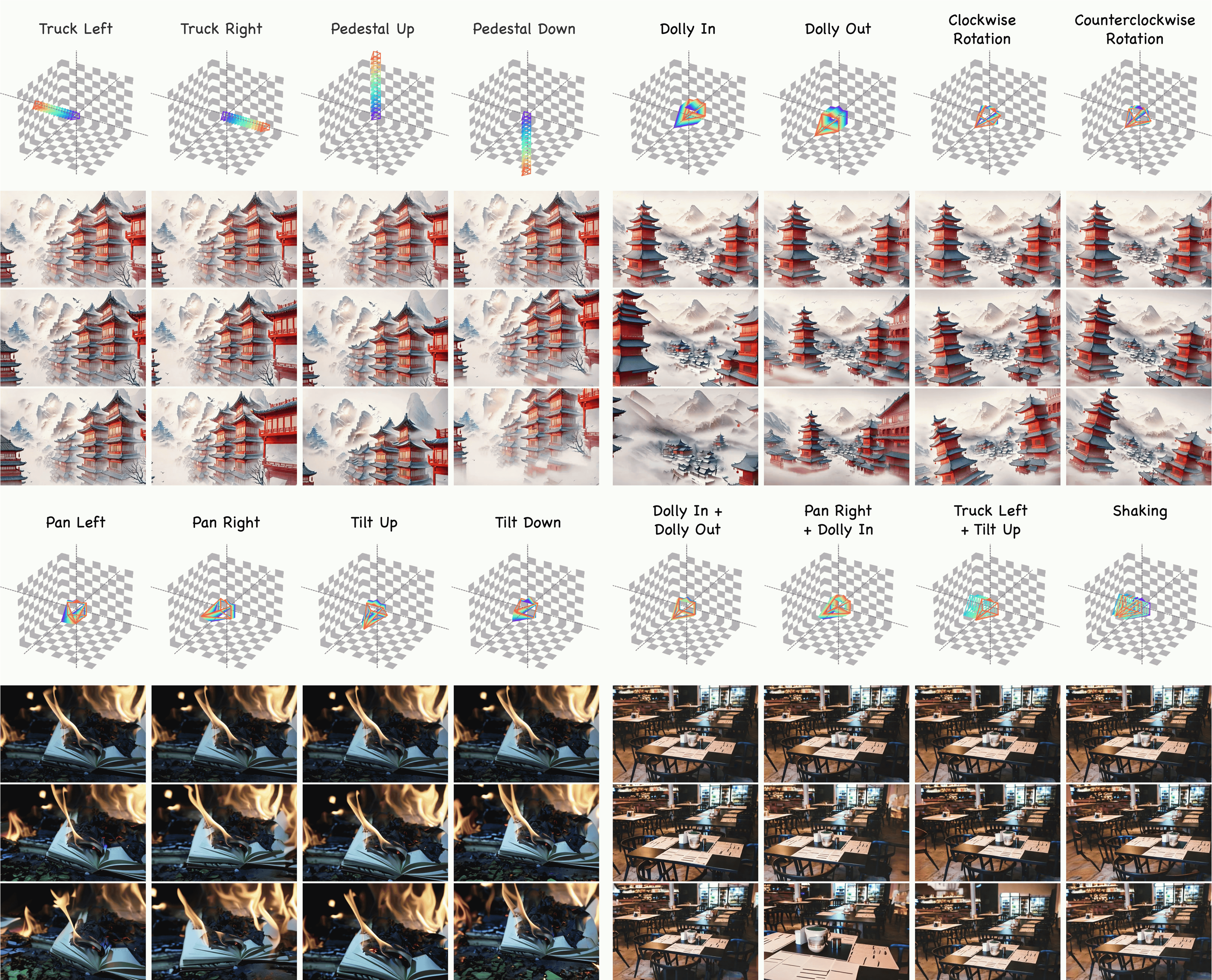}
    \caption{Results of camera-only motion control using our Perception-as-Control.}
    \label{fig:camera-only}
\end{figure*}

\begin{figure}[!ht]
    \centering
    \includegraphics[width=1.0\linewidth]{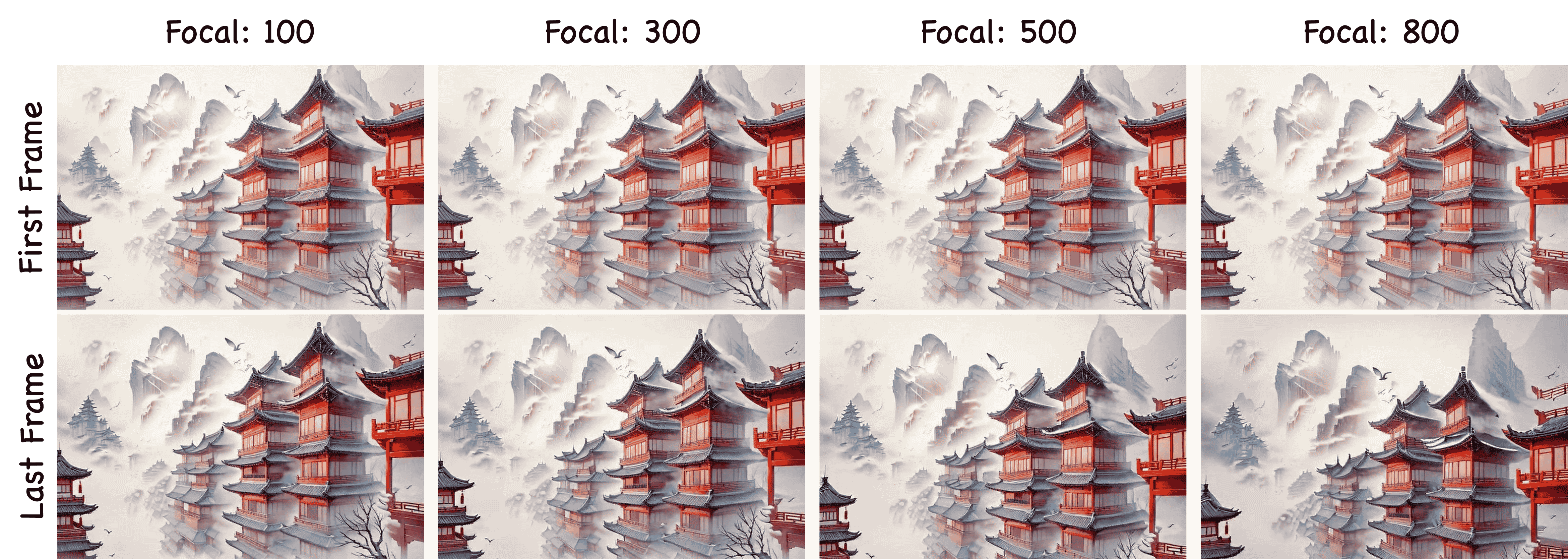}
    \caption{Results with the same camera extrinsic sequence (Pedestal Up) under different focal lengths.}
    \label{fig:focal_study}
\end{figure}

\section*{A. More Results of Perception-as-Control}

Perception-as-Control can achieve precise camera motion control adhering to any designated camera trajectory, multi-instance object motion control with adaptive control granularity, and harmonious collaborative motion control.

\begin{figure*}[!t]
    \centering
    \includegraphics[width=1.0\linewidth]{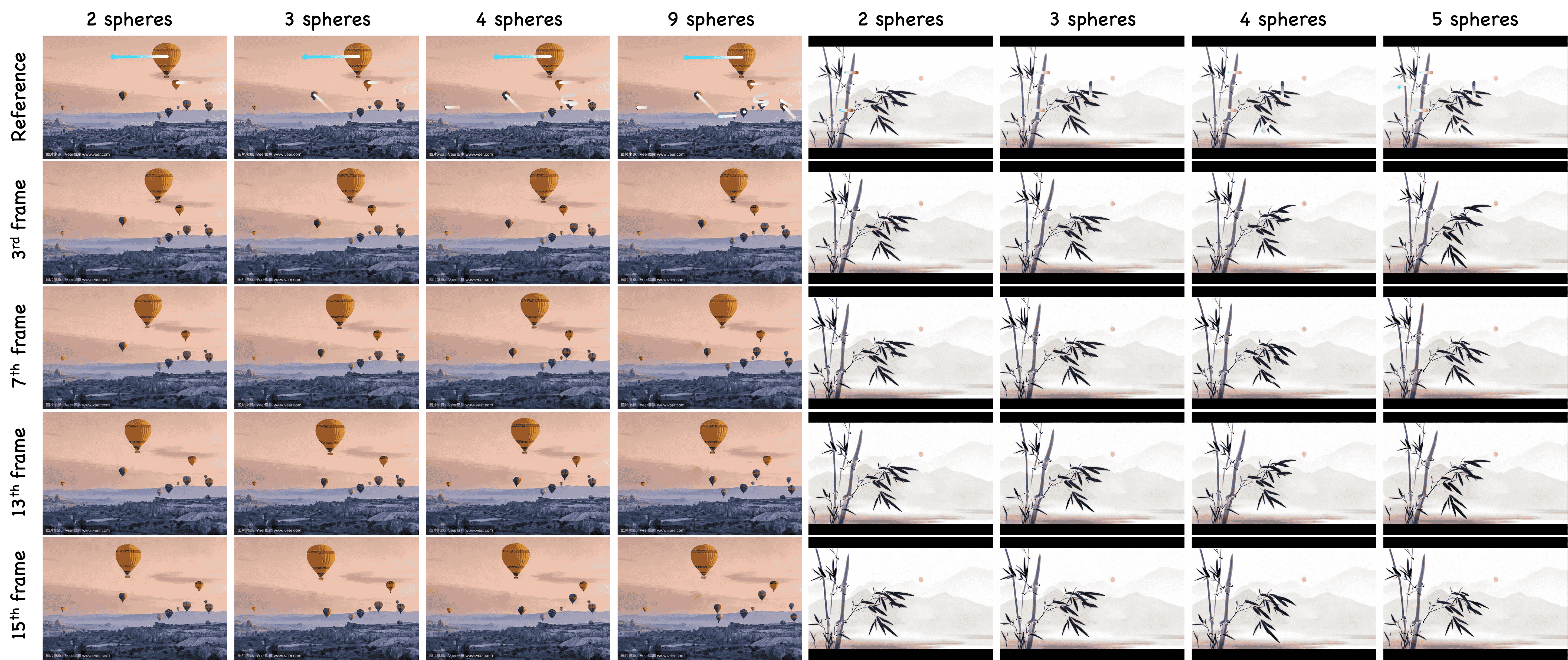}
    \caption{Results of object-only motion control using our Perception-as-Control.}
    \label{fig:object-only}
\end{figure*}

\subsection*{A.1. Camera-only Motion Control}

To demonstrate the superiority of Perception-as-Control in camera-only motion control, we show the generated results under different camera trajectories in Fig.~\ref{fig:camera-only}.
Our approach places no restrictions on the camera trajectories and supports camera motion control under basic camera trajectories such as Pan Left and Dolly In, their combinations such as Pan Left + Zoom In, and any user-provided camera trajectories.
Please refer to our \href{https://chen-yingjie.github.io/projects/Perception-as-Control}{webpage}: \emph{Fine-grained Collaborative Motion Control - Camera-only Motion Control} for the generated videos.

As shown in Fig.~\ref{fig:camera-only}, our framework can generate videos that adhere to any designated camera trajectories.
The generated videos maintain stable and consistent spatial structures owing to the world envelope in our 3D-aware motion representation, which demonstrates the direction and magnitude of camera movements and enhances overall perception.
Moreover, by adding several unit spheres to essential object parts, the generated videos are aware of depth information and can accurately reflect perspective effects under different focal lengths, as shown in Fig.~\ref{fig:focal_study}.

\subsection*{A.2. Object-only Motion Control}

For object-only motion control, we show the generated results in Fig.~\ref{fig:object-only} to demonstrate the superiority of our approach in achieving flexible and precise multi-instance object motion control with adaptive control granularity.
Please refer to our \href{https://chen-yingjie.github.io/projects/Perception-as-Control}{webpage}: \emph{Fine-grained Collaborative Motion Control - Object-only Motion Control}.

For visualization clarity, we project the unit sphere centroids onto the reference image to show how they change over time, using colors to represent the direction of movement.
As shown in the left part of Fig.~\ref{fig:object-only}, we control an increasing number of hot air balloons in the reference image, and the generated videos precisely reflect the movements of each one. In the right part of Fig.~\ref{fig:object-only}, when additional controls are applied to the bamboo, the proposed method can adaptively determine the control range for each unit sphere and achieve harmonious collaborative control results.

\begin{figure*}[!ht]
    \centering
    \includegraphics[width=1.0\linewidth]{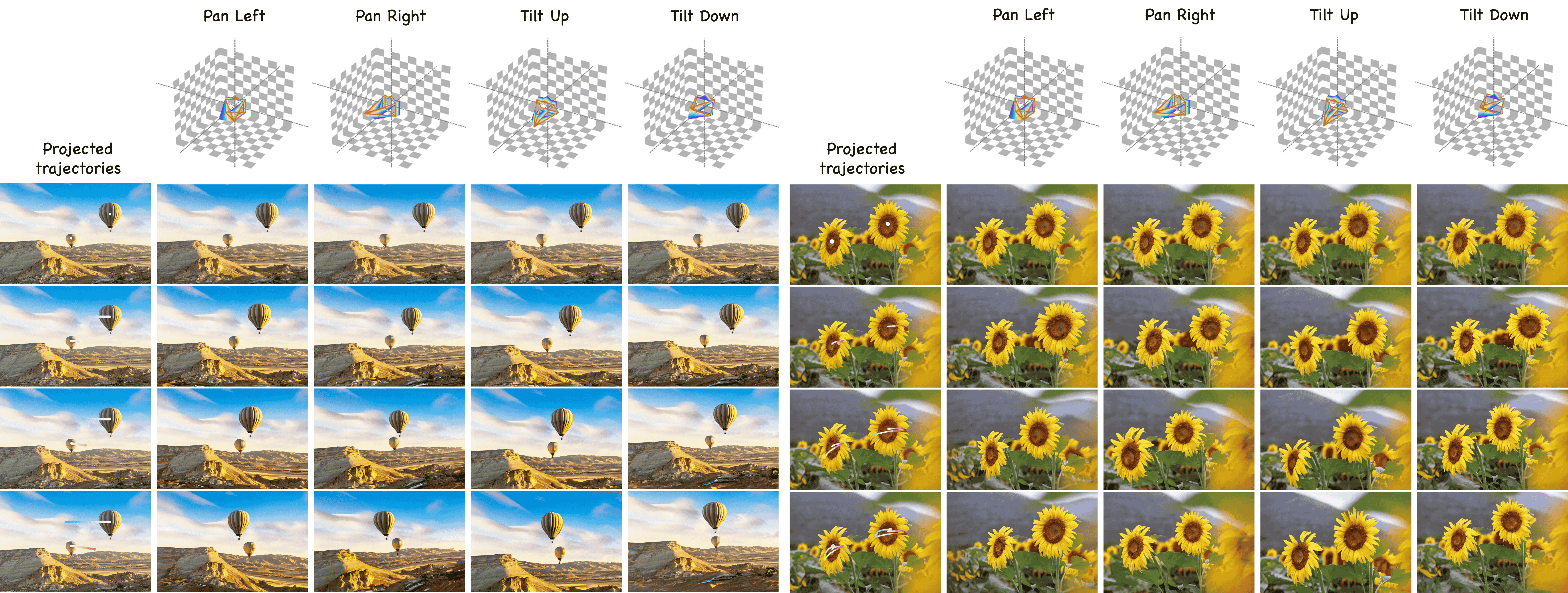}
    \caption{Results of collaborative motion control.}
    \label{fig:collaborative}
\end{figure*}

\subsection*{A.3. Collaborative Motion Control}
Our framework also supports collaborative control of both camera and object motions. 
Please refer to our \href{https://chen-yingjie.github.io/projects/Perception-as-Control}{webpage}: \emph{Fine-grained Collaborative Motion Control - Collaborative Motion Control} for the generated videos.

As illustrated in Fig.~\ref{fig:collaborative}, we control both camera and object motions and visualize the corresponding 3D-aware motion representation.
Our method can handle both large object motions and delicate object motions. The generated videos adhere to the specified camera motion and accurately present the adapted object motions.

\section*{B. More Details of ++ Versions}

\begin{figure*}[!ht]
    \centering
    \includegraphics[width=1.0\linewidth]{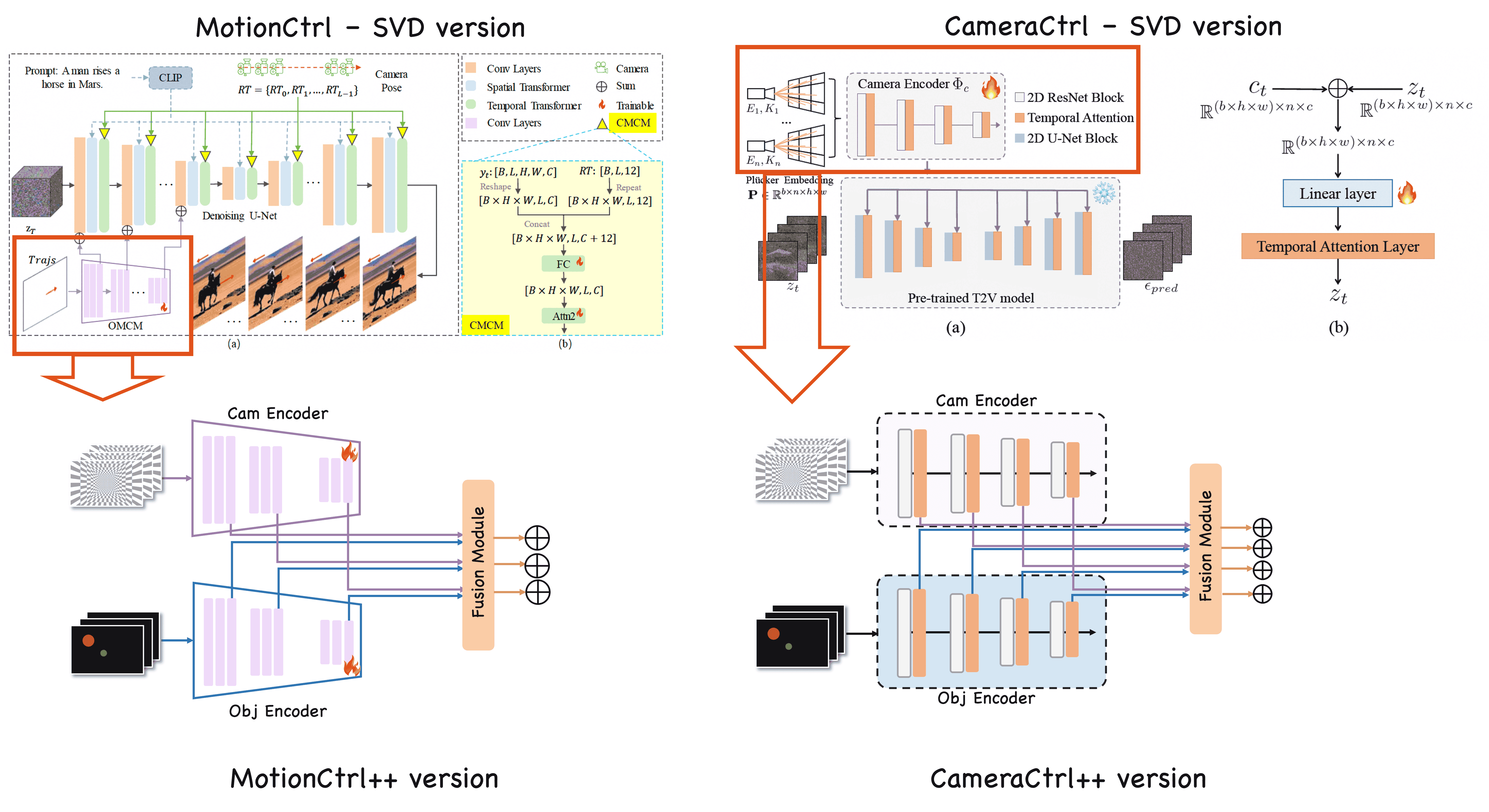}
    \caption{The frameworks of MotionCtrl++ and CameraCtrl++.}
    \label{fig:++version}
\end{figure*}

To demonstrate that our 3D-aware motion representation is orthogonal to base model architectures and effective for different base model architectures, we incorporate 3D-aware motion representation into the SVD~\cite{blattmann2023stable} version of CameraCtrl~\cite{he2024cameractrl} and MotionCtrl~\cite{wang2024motionctrl} to empower them with fine-grained collaborative motion control capability. Without making major changes to their architectures, we successfully incorporate our motion representation into them, as shown in Fig.~\ref{fig:++version}. The experimental results verify the effectiveness of our 3D-aware motion representation and the superiority of our framework.

\section*{C. More Details of Ablation Study}

\begin{figure*}[!ht]
    \centering
    \includegraphics[width=1.0\linewidth]{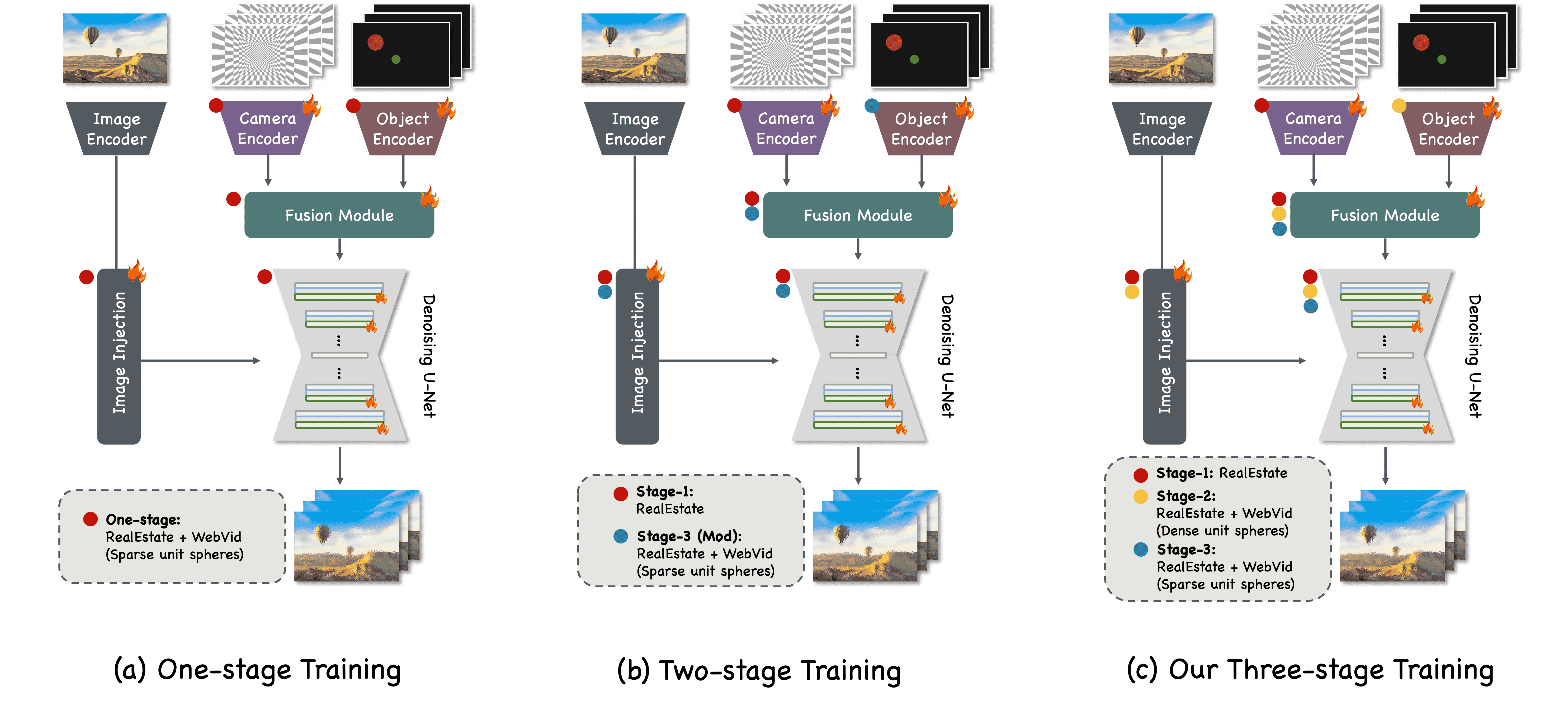}
    \caption{Illustration of different training strategies in ablation study.}
    \label{fig:training_strategy}
\end{figure*}

In Fig.~\ref{fig:training_strategy}, we illustrate the three training strategies included in our ablation study (Table 4). \emph{One-stage} refers to directly optimizing all the modules mentioned using mixed data and \emph{Two-stage} refers to merging the last two stages into one. 
 
\section*{D. More Details of Applications}

\begin{figure*}[!t]
    \centering
    \includegraphics[width=1.0\linewidth]{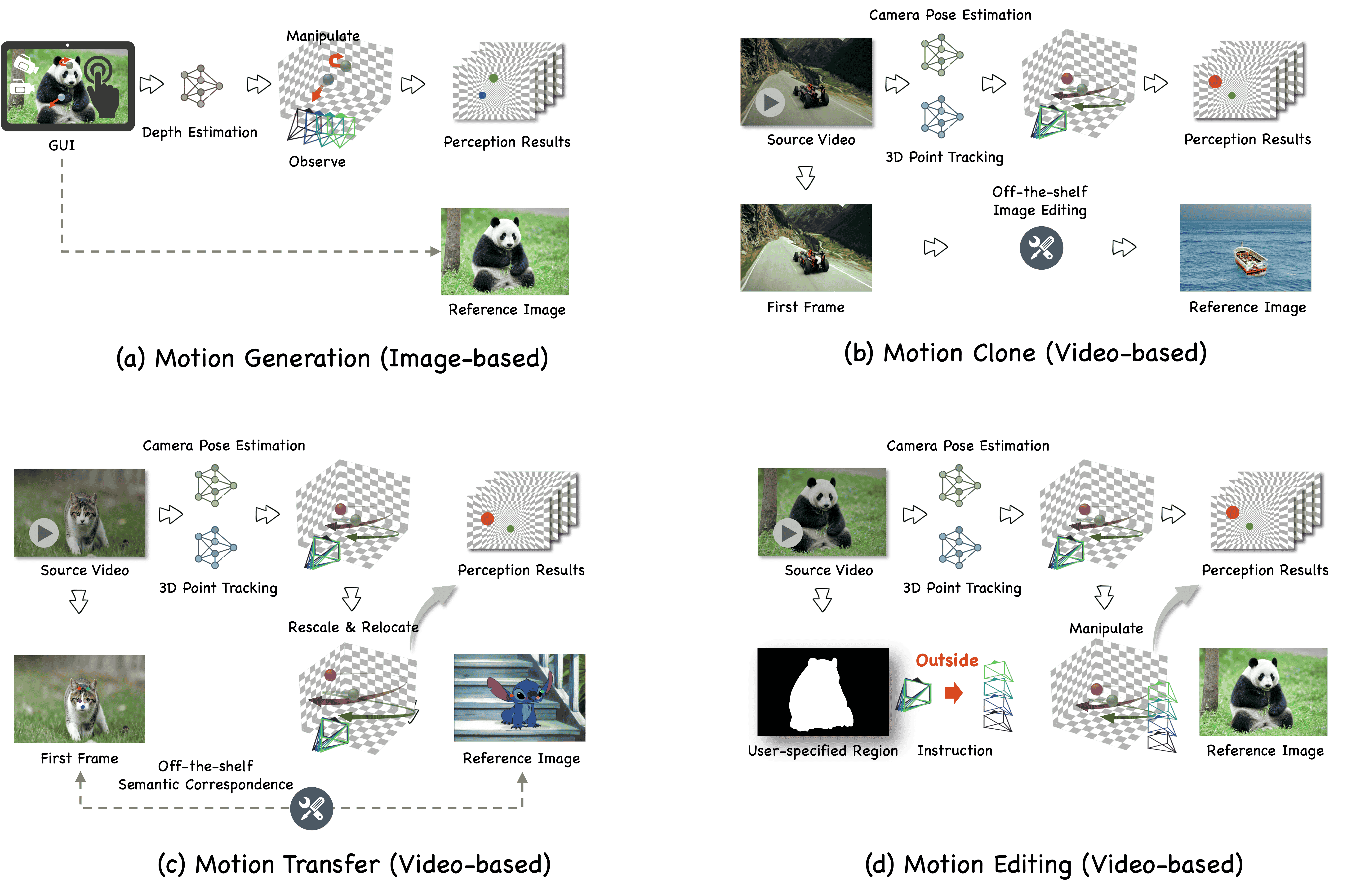}
    \caption{Illustration of different motion-related applications.}
    \label{fig:application_definitions}
\end{figure*}

By carefully interpreting user intentions into manipulation on 3D-aware motion representations, our framework can support various potential motion-related applications.
In this paper, we describe four kinds of motion-related applications, including motion generation, motion clone, motion transfer, and motion editing. 
The intuitive illustration is shown in Fig.~\ref{fig:application_definitions}.
Please refer to our \href{https://chen-yingjie.github.io/projects/Perception-as-Control}{webpage}: \emph{Potential Applications} for the generated videos.

\textbf{Motion Generation} refers to the task setting of interactive image animation that allows users to draw 2D or 3D point trajectories on the reference image and set an arbitrary camera trajectory with a carefully designed user interface. 
In this application, we employ an off-the-shelf metric depth estimation algorithm on the reference image, construct our 3D-aware motion representation based on the estimated results. We manipulate unit spheres corresponding to the starting point of each trajectory, and perceive the representation based on user-designated camera trajectory. The perception results are used as motion control signals.
Ideally, the object parts corresponding to the starting point of each trajectory should move along the trajectory in the generated video.
The difficulty lies in correctly understanding the control range of each trajectory point and generating reasonable and accurate motion videos under the control of multiple trajectories.

\textbf{Motion Clone} refers to extracting all motions from a source video and cloning them for driving a reference image. Users begin by providing a source video, which is used to construct our 3D-aware motion representation. They then edit the first frame of the source video to create a reference image. After that, we use the constructed 3D-aware motion representation and the reference image as inputs of the proposed framework to obtain the motion clone results.

\textbf{Motion Transfer}, similar to motion clone, refers to transferring local object motions from the source video to the reference image. In this case, the object may differ from that in the source video in terms of scale and location. 
To meet the needs of this application, we first find the semantic correspondences between the first frame of the source video and the reference image. Then, we construct our 3D-aware motion representation from the source video and locate essential unit spheres based on semantic correspondence points on the first frame. after that, we adaptively relocate and rescale them to match those semantic correspondence points on the reference image. Finally, the manipulated 3D-aware motion representation can be used to animate the reference image.

\textbf{Motion Editing} has greater control freedom than the above settings. Based on their source video, users provide segmentation masks and instructions to edit motions inside or outside these masks, \ie, editing fine-grained local motions in user-specified regions. In this way, users could freeze or modify motions of several objects or the background, \etc. To achieve such operations, we first construct 3D-aware motion representation from the source video, and select unit spheres inside the regions to be edited. Then, we manipulate those selected unit spheres according to user intentions and use the manipulated representation to re-animate the first frame of the source video.

\begin{figure}[!ht]
    \centering
    \includegraphics[width=1.0\linewidth]{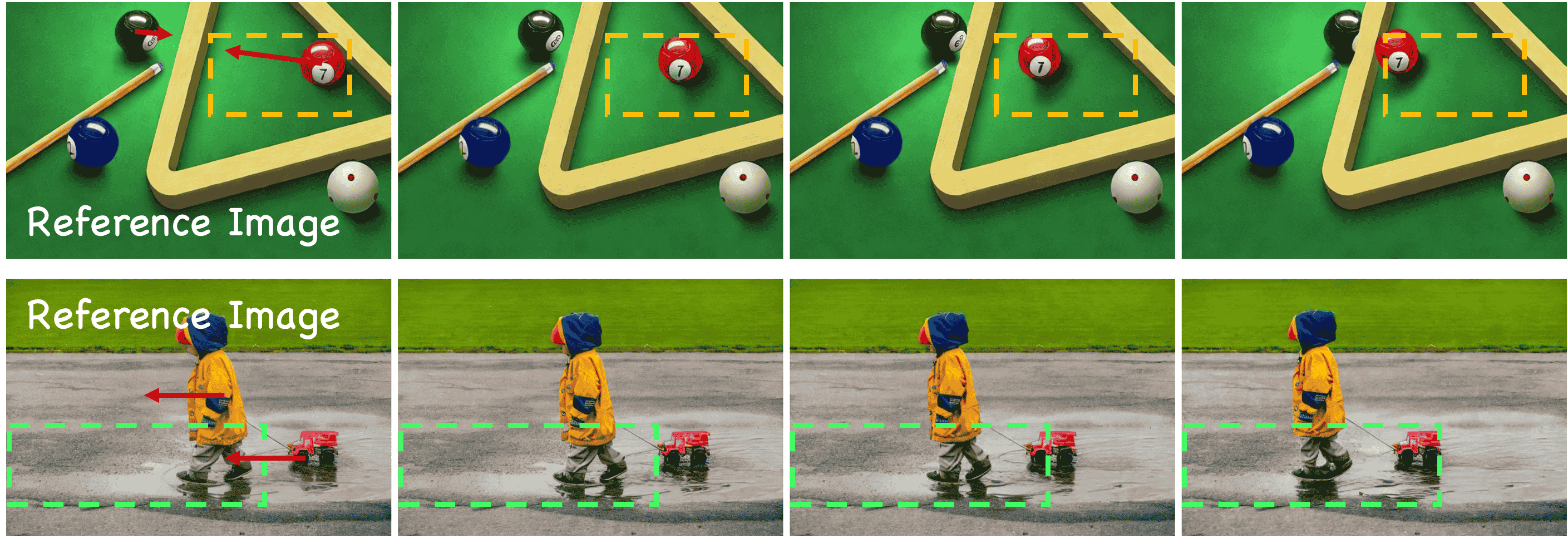}
    \caption{Limitations of Perception-as-Control.}
    \label{fig:limitations}
\end{figure}

\section*{E. Limitations and Future Work}
Our work introduces a novel 3D-aware motion representation for motion-controllable image animation task. Although our image animation framework that incorporates 3D-aware motion representation achieves promising and insightful results in fine-grained collaborative motion control, it still has some limitations. 

First, we simplify object parts as unit spheres, which can hardly reflect object rotation. As shown in the first row of Fig.~\ref{fig:limitations}, the relative positions between the two billiards and the markers on them remain unchanged, \ie, those billiards only move in translation without rotation. In future work, exploring more powerful alternatives may help express more comprehensive forms of movement.

Second, our approach is constrained by base model performance and cannot handle human-related motions satisfactorily. As shown in the second row of Fig.~\ref{fig:limitations}, the boy is moving forward without reasonable stepping movements. It is promising to incorporate the proposed 3D-aware motion representation into more powerful base model architectures.

Also, those off-the-shelf algorithms for constructing our 3D-aware representations influence the performance of our approach. Future work could employ state-of-the-art deep visual SLAM works for more precise estimation.


\end{document}